\documentclass[utf8]{FrontiersinHarvard} % for articles in journals using the Harvard Referencing Style (Author-Date)
\usepackage{url,hyperref,lineno,microtype,subcaption}
\usepackage{setspace}
\usepackage{tcolorbox}
\usepackage{multirow}
\usepackage{booktabs}
\usepackage[table]{xcolor}
\usepackage{tabularx}
\usepackage{array}
\usepackage{makecell}
\usepackage{amsmath}
\usepackage{float}
\makeatletter
\@removefromreset{equation}{section}
\makeatother

\usepackage{amssymb}
\usepackage{graphicx}
\newcolumntype{L}[1]{>{\raggedright\arraybackslash}p{#1}}
\newcolumntype{R}[1]{>{\raggedleft\arraybackslash}p{#1}}
\definecolor{tableheadgray}{RGB}{120,120,120}
\definecolor{tablelinegray}{RGB}{210,210,210}

\usepackage{soul}
\sethlcolor{yellow}      
\setul{0.5ex}{0.3ex}     
\setulcolor{yellow}      
\soulregister\citep7
\soulregister\citet7
\soulregister\ref7
\soulregister\textbf7

\newcolumntype{Y}{>{\centering\arraybackslash}X}

\newcommand{\safeincludegraphics}[2][]{%
  \IfFileExists{#2}{\includegraphics[#1]{#2}}{%
    \IfFileExists{#2.png}{\includegraphics[#1]{#2.png}}{%
      \IfFileExists{#2.pdf}{\includegraphics[#1]{#2.pdf}}{%
        \fbox{\parbox[c][0.22\textheight][c]{0.9\linewidth}{\centering Placeholder for #2}}%
      }%
    }%
  }%
}

\def\keyFont{\fontsize{8}{11}\helveticabold }
\def\firstAuthorLast{Yang {et~al.}}
\def\Authors{Xiaoli Yang\,$^{1,2}$, Huiyuan Tian\,$^{1}$, Yurui Li\,$^{1}$, Jianyu Zhang\,$^{1}$, Shijian Li\,$^{1,2,*}$ and Gang Pan\,$^{1,2}$}
\def\Address{$^{1}$College of Computer Science and Technology, Zhejiang University, Hangzhou, China\\
$^{2}$State Key Lab of Brain-Machine Intelligence, Zhejiang University, Hangzhou, China}
\def\corrAuthor{Shijian Li}
\def\corrEmail{shijianli@zju.edu.cn}

\begin{document}
\singlespacing
\onecolumn
\firstpage{1}

\title[Brain-CLIPLM: Semantic Compression for EEG-to-Text Decoding]{Brain-CLIPLM: Semantic Compression for EEG-to-Text Decoding} 

\author[\firstAuthorLast ]{\Authors}
\address{}
\correspondance{}
\extraAuth{}

\maketitle

{\noindent\small\textbf{Word count (main text):} 8,315 \quad \textbf{Figures:} 5 \quad \textbf{Tables:} 4\par}

\begin{abstract}
Decoding natural language from non-invasive electroencephalography (EEG) remains constrained by low signal-to-noise ratio and limited information bandwidth. This raises a central question: can sentence-level language be reliably recovered from such signals? Under realistic information constraints, this direct-recovery assumption may be too strong. We introduce a semantic compression hypothesis: non-invasive EEG may preserve recoverable semantic anchors rather than the full lexical--syntactic form of a sentence. From this perspective, direct sentence reconstruction is overly fine-grained relative to the recoverable information scale of EEG. To address this mismatch, we propose Brain-CLIPLM, a two-stage framework that decomposes EEG-to-text decoding into semantic-anchor recovery and anchor-guided sentence reconstruction. Stage 1 uses contrastive learning to align word-level EEG evidence with a fixed keyword vocabulary and recover ordered semantic anchors. Stage 2 uses a retrieval-grounded large language model with chain-of-thought reasoning prompts to reconstruct sentence meaning from these anchors, following a granularity matching principle that aligns decoding complexity with the recoverable neural information scale. On the combined Zurich Cognitive Language Processing (ZuCo) benchmark, Brain-CLIPLM achieves 67.6\% Top-5 and 85.0\% Top-25 sentence retrieval accuracy, with the strongest performance at intermediate anchor granularity. Control analyses show that EEG-derived anchors carry sentence-specific information beyond language-model priors. Within the constrained ZuCo sentence pool and fixed keyword-vocabulary settings, these findings suggest that EEG-to-text decoding is better framed as recovering compressed semantic content before anchor-guided sentence reconstruction.

\tiny
\keyFont{\section{Keywords:} EEG-to-text decoding, semantic anchors, semantic compression, sentence reconstruction, large language models, contrastive learning, brain-computer interface}

\end{abstract}

\section{Introduction}

After reading a sentence, people often retain its core meaning more reliably than its exact wording. Human sentence comprehension does not require a verbatim recovery of every word. Classic studies of sentence memory and text comprehension show that readers preserve semantic gist more robustly than surface form, and can often reconstruct sentence meaning even when exact wording is no longer available \citep{sachs1967recognition,kintsch1978toward,potter1990regeneration}. Neurocognitive studies further suggest that lexical access, semantic composition, and sentence-level interpretation are related but separable components of language comprehension \citep{friederici2011brain,pylkkanen2019neural}. This cognitive property carries an important implication for EEG-based language decoding: the most recoverable part of a sentence may not be its complete lexical--syntactic form, but a smaller set of semantic cues that constrain its core meaning. This distinction matters not only for modeling language comprehension, but also for building practical EEG-based communication systems.

\begin{figure}[htbp]
\centering
\includegraphics[width=\textwidth]{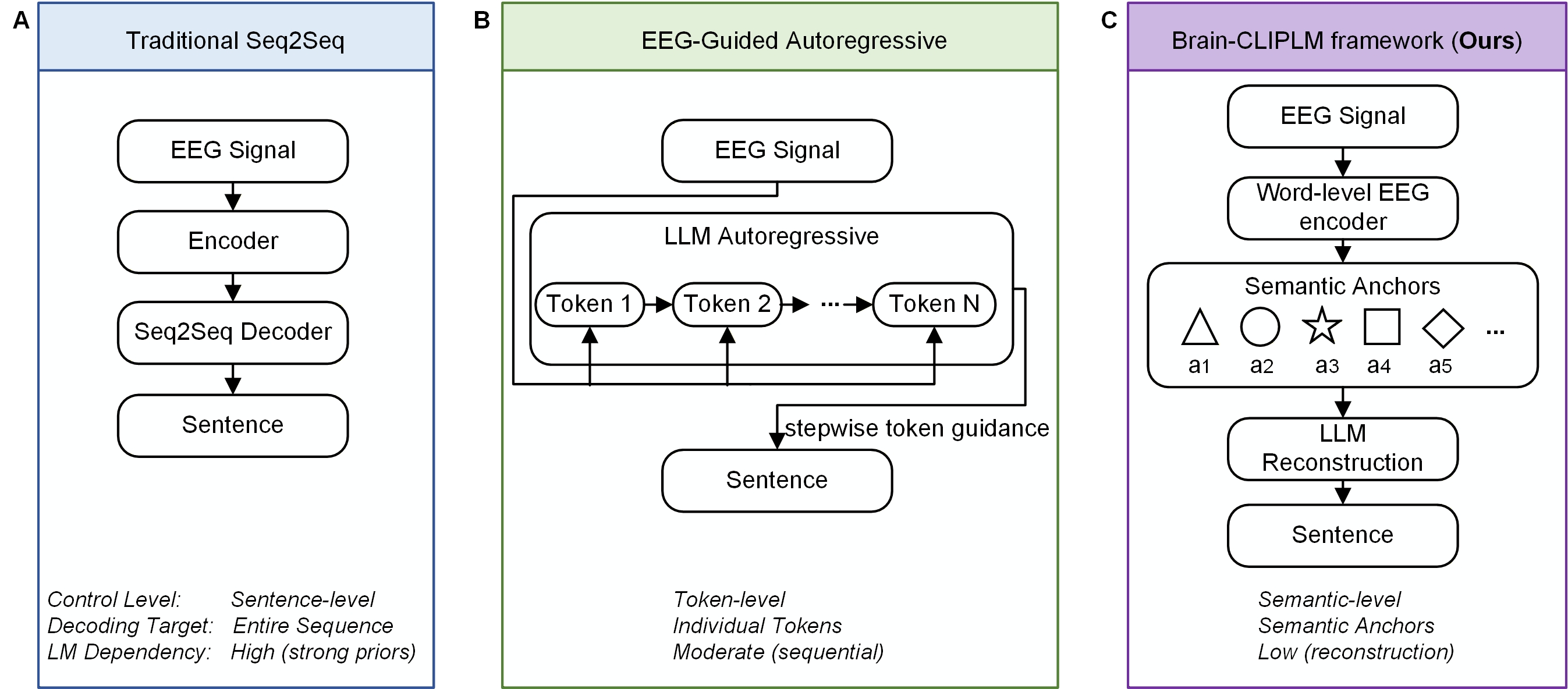}
\caption{Conceptual comparison of three EEG-to-text decoding paradigms. \textbf{(A)} Traditional sequence-to-sequence decoding maps EEG signals directly to sentence output. \textbf{(B)} EEG-guided autoregressive decoding uses neural representations to guide token-level candidate generation. \textbf{(C)} Brain-CLIPLM factorizes EEG-to-text decoding into EEG-to-keyword recovery followed by keyword-to-sentence reconstruction, treating recovered keywords as ordered semantic anchors for sentence reconstruction.}
\label{figure1}
\end{figure}

Yet many EEG-to-text formulations still place the decoding target at the sentence level. Existing studies can be broadly grouped into two paradigms. The first follows a traditional sequence-to-sequence formulation, in which an EEG encoder is paired with a text decoder to map neural signals directly to sentence output \citep{wang2022open, duan2023dewave, liu2024eeg2text}. The second uses EEG-derived representations to guide autoregressive language generation, typically by influencing token-level candidate selection or generation \citep{ZHENG2026130300}. These two paradigms are summarized in Figure~\ref{figure1}A,B. Both paradigms, however, still place a fine-grained linguistic burden on EEG. The sequence-to-sequence route asks a single non-invasive signal---affected by artifacts, cross-subject variability, and limited bandwidth \citep{bertrand1985average,jung2000removing,nieto2022thinking}---to support direct sentence-level recovery. At the same time, prior studies on speech and language processing suggest that neural oscillatory activity carries information at multiple temporal scales, providing a rationale for analyzing EEG through frequency-filtered features. Lower-frequency dynamics, including delta- and theta-band responses, have been linked to cortical tracking of slower linguistic units and speech comprehension, whereas beta/gamma-band activity has been implicated in predictive and higher-level language processing \citep{giraud2012cortical,ding2016cortical,etard2019neural,lewis2015predictive}. In this study, such frequency-filtered EEG features are used as an engineering representation of language-related neural evidence, rather than as a claim that any single frequency band uniquely encodes semantic anchors or sentence meaning. The token-guided autoregressive route relaxes this burden by using an LLM to generate candidates, but it still requires EEG to provide stepwise guidance at the token level, where errors can accumulate across the sentence. These targets remain more detailed than the compact semantic evidence that non-invasive EEG is likely to support. They also demand large amounts of EEG--sentence paired data, which are scarce and expensive to collect. When neural evidence is weak, a powerful language model may also fill in fluent text from its own priors rather than from sentence-specific EEG information \citep{wang2022open, jo2024eeg, jo2025evaluating}. These issues point to a simpler question: instead of asking EEG to directly generate a whole sentence, can it first recover the semantic anchors that make sentence reconstruction possible?

We answer this question by treating EEG as partial semantic evidence rather than a complete transcript. We propose Brain-CLIPLM, a semantic-factorized EEG-to-text decoding framework that combines contrastive EEG-to-keyword alignment with language-model-based sentence reconstruction. As illustrated in Figure~\ref{figure1}C, it factorizes EEG-to-text decoding into ordered semantic-anchor recovery followed by anchor-guided sentence reconstruction. In the first stage, fixation-aligned word-level EEG segments are aligned with a fixed keyword vocabulary, and the decoded keywords inherit the original reading order as ordered semantic anchors. The output is therefore not an unordered keyword bag, nor a full word sequence, but an ordered semantic scaffold of the sentence. In the second stage, a large language model reconstructs sentence meaning from this scaffold, supported by retrieval from the task-specific corpus and chain-of-thought (CoT) reasoning prompts. This factorization separates neural semantic recovery, which EEG can plausibly support, from linguistic surface realization, which a language model is better suited to perform. The semantic compression hypothesis behind this design is deliberately conservative: we do not claim that EEG contains full sentence form in compressed code, but that EEG may support a compact, ordered set of semantic cues that makes constrained sentence recovery more tractable. Importantly, the use of multiple anchors, rather than a single one, reflects the fact that sentence meaning is distributed over a small number of semantically coherent units, a view broadly consistent with evidence that online prediction is constrained by constituent-level structure \citep{zou2026constituent}. This makes anchor-level decoding both structurally plausible and practically effective as an intermediate target.

The present study makes three main contributions. First, we propose a conceptual framework for EEG language decoding grounded in the semantic compression hypothesis and information-theoretic constraints, offering a perspective on why anchor-based decoding may provide a better-matched target for non-invasive EEG than direct sentence reconstruction under the present constraints. Second, we develop Brain-CLIPLM as a concrete implementation of this framework, combining contrastive neural representation learning with large language model-based sentence reconstruction. Third, we provide comprehensive empirical validation, including anchor-granularity analyses, the control-baseline comparison of anchor conditions, reconstruction ablations, and cross-subject experiments on the ZuCo benchmark, showing that EEG-derived semantic representations capture sentence-specific information beyond language-model priors. Together, these contributions support framing EEG-to-text decoding as the recovery of compressed semantic content rather than the direct reconstruction of complete linguistic form.

The remainder of this paper is organized as follows. Section~2 develops the semantic compression hypothesis and explains why semantic anchors provide a more suitable intermediate target than direct full-sentence decoding. Section~3 describes the proposed Brain-CLIPLM framework, including the ZuCo benchmark, fixed keyword vocabulary construction, EEG-to-keyword alignment, and the anchor-guided sentence reconstruction pipeline. Section~4 reports the main experimental results, including target-granularity analyses, semantic-anchor decoding, sentence reconstruction, control comparisons, and ablation studies. Section~5 discusses the implications and limitations of Brain-CLIPLM. Finally, Section~6 concludes the paper.

\section{Semantic Compression as the Theoretical Basis of Brain-CLIPLM}

\begin{figure}[htbp]
\centering
\includegraphics[width=\textwidth]{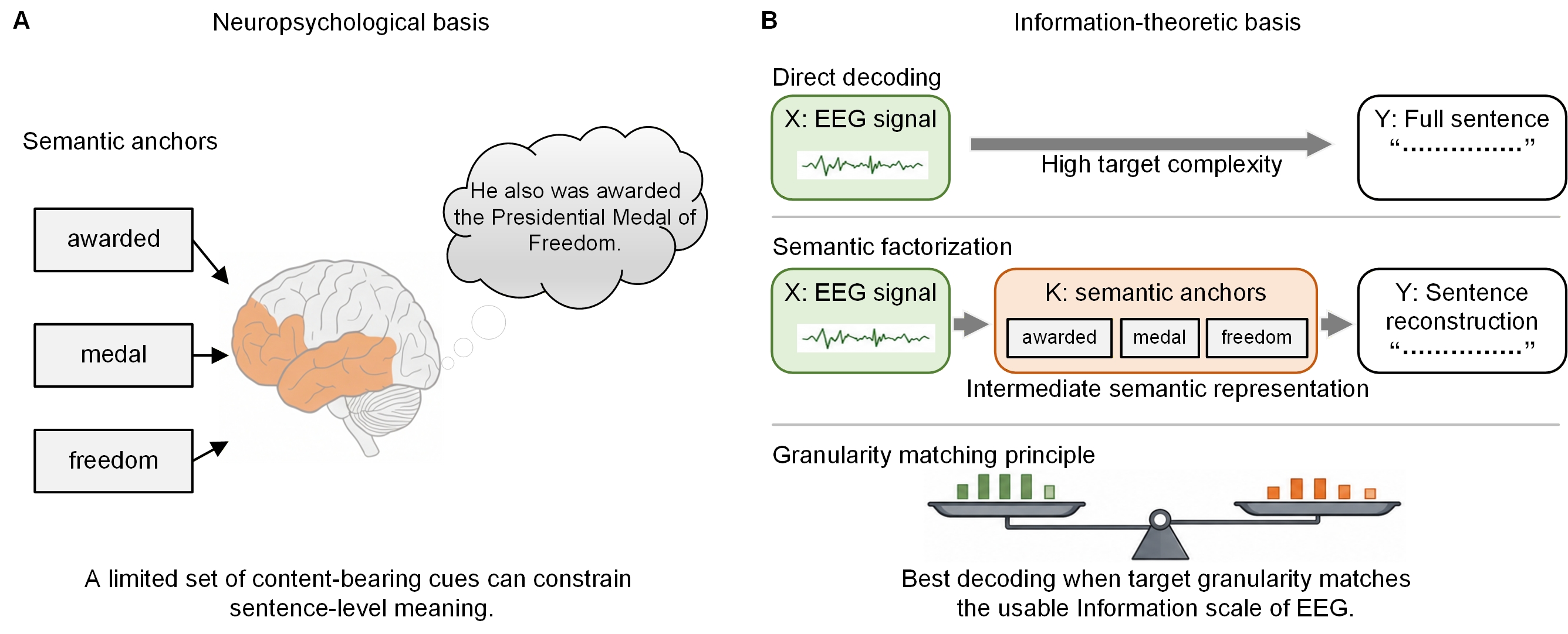}
\caption{Neuropsychological and information-theoretic foundations of the semantic compression hypothesis. \textbf{(A)} Neuropsychological basis: a limited set of content-bearing semantic anchors can support sentence-level meaning. \textbf{(B)} Information-theoretic basis: an intermediate semantic-anchor representation provides a better-matched decoding target for non-invasive EEG than direct full-sentence decoding.}
\label{figure2}
\end{figure}

The central question for Brain-CLIPLM is whether non-invasive EEG should be asked to recover a complete sentence directly, or whether it should first be mapped to a more compact semantic representation. This section provides the theoretical basis for the design and evaluation of Brain-CLIPLM. It develops the semantic compression hypothesis from two complementary perspectives. The first is cognitive: sentence meaning can often be supported by a limited set of content-bearing cues. The second is information-theoretic: recovering such cues is a more suitable target for noisy EEG than specifying a full lexical--syntactic sentence form. Together, these perspectives explain why Brain-CLIPLM first recovers semantic anchors and only then reconstructs the sentence. Figure~\ref{figure2} summarizes the two foundations of the semantic compression hypothesis. Figure~\ref{figure2}A illustrates the neuropsychological basis: a limited set of semantic anchors can support sentence-level meaning. Figure~\ref{figure2}B illustrates the information-theoretic basis: an intermediate semantic representation provides a better-matched target for non-invasive EEG than direct full-sentence decoding.

In this study, the semantic compression hypothesis and the Brain-CLIPLM framework play different but connected roles. The semantic compression hypothesis is the theoretical claim: non-invasive EEG may not reliably support direct recovery of the full lexical--syntactic form of a sentence, but it may preserve enough semantic evidence to recover an ordered set of content-bearing anchors. The granularity matching principle then specifies the modeling implication of this claim: the decoding target should match the recoverable information scale of non-invasive EEG. Brain-CLIPLM operationalizes this theoretical claim as a testable decoding pathway. It implements the factorized form $X \rightarrow K \rightarrow Y$, where $X$ denotes EEG evidence, $K$ denotes the recovered semantic-anchor representation, and $Y$ denotes the reconstructed sentence. In this formulation, the semantic compression hypothesis defines what should be decoded from EEG, whereas Brain-CLIPLM specifies how that target can be recovered and evaluated. Specifically, semantic anchors refer to EEG-recovered content-bearing decoded keywords that serve as an intermediate semantic representation and as ordered constraints for sentence reconstruction. They are not ordinary text keywords or sentence-level tokens. The fixed keyword vocabulary defines the candidate space for Stage 1, whereas semantic anchors denote the recovered keywords when they are used to constrain Stage 2 reconstruction.

\subsection{Neuropsychological basis}

The neuropsychological motivation for semantic anchors starts with a simple observation: sentence meaning is not carried uniformly by every word. Nouns, verbs, and adjectives often provide the main propositional content, while function words and syntax shape how that content is expressed. This does not make grammar, word order, or surface form unimportant. It means that the gist of a sentence can often remain accessible even when exact wording is incomplete or partially degraded. Classic studies of sentence memory and discourse comprehension support this distinction: people preserve meaning more robustly than verbatim form \citep{sachs1967recognition,kintsch1978toward,potter1990regeneration}. Neurocognitive accounts likewise separate semantic processing from the finer-grained sequencing and syntactic operations required for full sentence realization \citep{friederici2011brain,pylkkanen2019neural,pulvermuller2018neurobiological}. The semantic compression hypothesis is therefore deliberately conservative: it does not claim that syntax is irrelevant, but that the signal most stably recoverable from non-invasive EEG may lie at an intermediate semantic scale rather than at the level of verbatim transcription.

A second question follows: if sentence meaning can be compressed, why use multiple anchors rather than a single cue? The reason is that sentence meaning is usually not concentrated in one word or one latent point. It is distributed across a small number of semantically coherent units. Structurally, language is organized into phrases and clauses rather than a flat sequence of independent tokens, and recent evidence suggests that online word prediction is constrained by constituent boundaries \citep{zou2026constituent}. Semantically, several content-bearing elements may be needed to cover different parts of a proposition. Effective compression should therefore preserve several informative components rather than collapse the entire sentence into a single cue.

This view clarifies the role of semantic anchors in Brain-CLIPLM. Anchors are not assumed to correspond one-to-one to parser-defined syntactic constituents, nor do they aim to reproduce full lexical detail. Instead, they provide a compact set of semantically informative elements that reflect the multi-unit nature of sentence meaning. In the present study, this idea is instantiated through the ZuCo natural reading corpus. Brain-CLIPLM defines its candidate space as a fixed keyword vocabulary drawn from the task-specific sentence corpus. This vocabulary is restricted to linguistically informative words, primarily nouns, verbs, and adjectives, and remains fixed within each task and vocabulary size. During Stage~1, fixation-aligned word-level EEG segments provide local neural evidence for these candidate keywords. During Stage~2, the recovered keywords are kept in their original reading order and treated as semantic anchors for sentence reconstruction. Thus, the claim is not that the brain stores sentences as keyword lists, or that anchors uniquely determine the sentence. The narrower claim is that, under non-invasive EEG constraints, recovering a small ordered set of semantic anchors is a more plausible first target than recovering a complete sentence.

\subsection{Information-theoretic basis}

The cognitive argument explains why semantic anchors are meaningful. The remaining question is whether they also provide a better-matched target for the information scale of non-invasive EEG. In information theory, entropy provides a natural way to describe the scale of an output space and the number of bits required to specify an outcome \citep{shannon1948mathematical,cover1999elements}. Let $X$ denote the EEG signal, $X_{\mathrm{sem}}$ the task-relevant semantic information that can be extracted from EEG, $K_m$ an ordered semantic-anchor representation of length $m$, and $Y$ the full sentence. We do not equate the raw dimensionality of $X$ with usable linguistic information, because EEG measurements are continuous, correlated across time and space, and contaminated by noise, artifacts, and volume-conducted mixing \citep{bertrand1985average,jung2000removing,nieto2022thinking}. The relevant comparison is instead between the recoverable semantic evidence in $X$, the anchor-level target $K_m$, and the full sentence target $Y$:
\[
X_{\mathrm{sem}} \text{ is evaluated at the scale of } H(K_m), \qquad H(K_m) \ll H(Y).
\]
This relation captures the core information-scale claim of Brain-CLIPLM: EEG is more plausibly used to recover a compact semantic scaffold than to directly specify the full lexical--syntactic form of a sentence.

In the primary Brain-CLIPLM setting, this comparison becomes concrete. The keyword vocabulary contains $V=100$ items, and the ordered anchor lengths are $m \in \{3,5,7\}$. The corresponding anchor-level targets are approximately 19.9, 33.2, and 46.5 bits, respectively. The contrast with full sentence recovery is substantial: even a deliberately conservative 20-word sentence built from the same 100-word lexical alphabet is already at least 132.9 bits. The real sentence space is larger still, because full sentences require function words, richer vocabulary, correct word order, local syntax, and fluent surface realization. These values are target-complexity proxies rather than estimates of the total information capacity of EEG. Their purpose is to show the scale gap between an anchor-level target and a full sentence target.

This comparison leads to the granularity matching principle. A single anchor is too coarse to capture distributed semantic content, whereas full sentence reconstruction is too fine-grained relative to the available neural signal. Semantic anchors occupy the intermediate level: structured enough to preserve meaning, but compact enough to remain a feasible EEG decoding target. The cognitive and information-scale arguments therefore point to the same conclusion. EEG should first be used to recover partial semantic evidence, and sentence reconstruction should be performed only after this evidence has been made explicit. In Brain-CLIPLM, Stage~1 solves this constrained neural decoding problem by estimating ordered semantic anchors from word-level EEG segments, whereas Stage~2 solves a constrained language realization problem by using these anchors to guide sentence reconstruction. This separation assigns semantic support to the neural signal and surface realization to the language model, while keeping the recovered semantic evidence explicit and testable.

\section{Materials and Methods}

This section translates the semantic-anchor hypothesis into a reproducible EEG-to-text decoding pipeline. Section~3.1 defines the ZuCo benchmark and preprocessing pipeline, showing how natural reading data provide word-aligned EEG evidence while the evaluation remains sentence-level. Section~3.2 constructs a fixed keyword vocabulary, which defines the candidate space for Stage~1 and supports later semantic-anchor recovery. Section~3.3 presents Brain-CLIPLM: Stage~1 recovers ordered anchors from EEG, and Stage~2 reconstructs sentences from those anchors using an LLM with retrieval and reasoning prompts. Section~3.4 defines the evaluation metrics, controls, ablations, and statistical tests used to determine whether recovering anchors first is more effective than directly decoding complete sentences.

\subsection{Dataset and preprocessing}

We evaluate Brain-CLIPLM on the Zurich Cognitive Language Processing (ZuCo) benchmark, constructed from the publicly available ZuCo 1.0 and ZuCo 2.0 corpora \citep{hollenstein2018zuco, hollenstein2020zuco}. The combined benchmark comprises 30 participants, including 12 from ZuCo 1.0 and 18 from ZuCo 2.0, recorded during natural reading tasks. Following recent protocol-aligned EEG-to-text studies \citep{ZHENG2026130300, 2024Enhancing}, we build the benchmark from four commonly used task subsets: sentiment reading from movie reviews (SR v1.0), normal reading of Wikipedia sentences (NR v1.0 and NR v2.0), and task-specific reading of Wikipedia sentences (TSR v1.0). At the task level, SR v1.0 contains 400 unique sentences and 4,533 samples, NR v1.0 contains 300 unique sentences and 3,343 samples, NR v2.0 contains 349 unique sentences and 4,547 samples, and TSR v1.0 contains 407 unique sentences and 5,579 samples. To remain comparable with \citet{ZHENG2026130300}, we do not merge all tasks indiscriminately. Instead, we use SR v1.0, NR v1.0, and NR v2.0 in the multi-subject setting to maximize subject diversity, whereas the single-subject and cross-subject settings use SR v1.0, NR v1.0, and TSR v1.0 so that participants are evaluated on matched reading materials.

EEG signals are recorded with a 128-channel EEG Geodesic Hydrocel system at 500~Hz with an online frequency range of 0.1--100~Hz. Nine EOG channels are used for artifact monitoring and removal, and channels located primarily on the neck and face are discarded, leaving 105 scalp EEG channels for subsequent analysis. EEG signals are re-referenced to the common average reference \citep{bertrand1985average}, bad channels are interpolated using spherical splines \citep{perrin1989spherical}, ocular artifacts are removed using independent component analysis (ICA) \citep{jung2000removing}, and baseline correction is applied using the 200~ms interval preceding sentence onset. Eye-tracking fixations are then used to divide EEG recordings into word-aligned segments, and each segment is represented by the concatenated statistical features extracted from eight frequency-band filters---theta1 (4--6~Hz), theta2 (6.5--8~Hz), alpha1 (8.5--10~Hz), alpha2 (10.5--13~Hz), beta1 (13.5--18~Hz), beta2 (18.5--30~Hz), gamma1 (30.5--40~Hz), and gamma2 (40--49.5~Hz)---yielding an 840-dimensional feature vector for each word position. This eight-band word-level EEG representation follows the feature-extraction setting described in the ZuCo datasets \citep{hollenstein2018zuco, hollenstein2020zuco} and reused in prior ZuCo-based EEG-to-text work \citep{ZHENG2026130300, 2024Enhancing, jo2024eeg}.

Because the present framework uses word-level neural evidence as the input to Stage~1, EEG samples are organized into word-aligned feature-vector sequences rather than a single sentence-level representation. This design keeps the granularity of the neural input close to the granularity of the decoding target and preserves the original order of the content-bearing units within each sentence. At the same time, the aim of Stage~1 is not to recover a full lexical transcript word by word, but to extract an ordered sequence of semantically informative content anchors that can support subsequent sentence reconstruction.

The word alignment is obtained from the synchronized eye-tracking fixation annotations provided by ZuCo. This fixation-defined segmentation is applied consistently to training, validation, and test trials before model fitting. During validation and testing, the model receives EEG feature vectors extracted from fixation-defined word-level segments, but it does not receive the ground-truth word identity or the ground-truth sentence label as input. Word identity is used only for supervised Stage 1 label construction during training and for evaluating recovered anchors after prediction. We therefore treat eye-tracking-based segmentation as part of the measurement protocol rather than as textual label leakage.

To maintain reliable EEG--text alignment, we follow Zheng et al. \citep{ZHENG2026130300} and filter out samples in which the number of word-aligned EEG segments is less than half the total number of words in the corresponding sentence. For single-subject and multi-subject experiments, data are partitioned by unique sentences into training (80\%), validation (10\%), and test (10\%) sets, ensuring that test sentences are not observed during training. For cross-subject evaluation, we adopt a leave-one-subject-out protocol such that all EEG trials from the held-out participant are excluded from model fitting; in each fold, one subject is used for testing, one additional subject for validation, and the remaining subjects for training. This construction follows \citet{ZHENG2026130300} and facilitates closer protocol-level comparison. All experiments use publicly available, de-identified data. The original ZuCo studies were approved by the Ethics Committee of the University of Zurich, and all participants provided written informed consent. In this way, the dataset organization supports the central design of the study: EEG is used to recover ordered word-level semantic evidence, while final evaluation asks whether that evidence improves sentence-level recovery.

\subsection{Keyword vocabulary construction}

The keyword vocabulary is the bridge between the original sentence and the reduced target that Stage~1 must decode. Its role is not to list every token in the corpus, but to define a fixed candidate keyword space that is meaningful enough to guide sentence reconstruction and compact enough to be learnable from noisy EEG. A candidate keyword is selected as a high-frequency content word that satisfies two requirements: it carries sentence-level meaning, and it appears often enough to support EEG--text alignment. Because word-aligned EEG segments follow the original reading order, decoded keywords naturally form an ordered sequence that serves as semantic anchors for Stage~2 reconstruction. Once constructed, the vocabulary is fixed for all Stage~1, Stage~2, and control experiments, so performance differences cannot be attributed to changes in the target space. This distinction is important: content words define the linguistic candidate pool, the fixed keyword vocabulary defines the supervised decoding space, and decoded keywords become semantic anchors only when they are recovered from EEG and used as ordered constraints in Stage 2.

We construct this vocabulary in three steps. First, we build a linguistically constrained candidate pool from the ZuCo sentences. All sentences are lowercased, stripped of punctuation, tokenized, part-of-speech tagged, and lemmatized under a shared normalization pipeline. POS tags are mapped to WordNet-compatible categories \citep{miller1995wordnet}, and we retain nouns, proper nouns, verbs, and adjectives as candidate content words. We remove numbers, tokens containing digits, month names and abbreviations, common temporal expressions, written numerals, ordinals, common quantificational words, generational suffixes, pure Roman numerals, stop words, residual pronoun-like forms, and very short proper-name abbreviations. Personal names identified by named-entity recognition are excluded by default, whereas non-person entities are retained only when they satisfy the same frequency, coverage, and diversity criteria as other candidate keywords. This step ensures that the keyword pool mainly contains words that can carry sentence meaning rather than words that mainly support grammatical surface form.

Second, we balance learnability and semantic diversity. A minimum corpus-frequency threshold of five occurrences removes extremely sparse words that are unlikely to provide stable EEG supervision. Because frequency alone can over-select near-duplicate words, each candidate is represented with a 300-dimensional GloVe embedding \citep{pennington2014glove}, and a diversity-aware core vocabulary is selected by greedy farthest-point sampling \citep{gonzalez1985clustering} using cosine distance. Repeated sentence-internal occurrences of the same lexical item are counted only once for sentence-level coverage, and derivationally related forms sharing the same lexical root are not retained together when they form highly overlapping semantic families. This step prevents the keyword vocabulary from collapsing into a small cluster of frequent but redundant words.

Third, we audit sentence-level coverage. For each target vocabulary size $V \in \{50,100,200,500\}$, we first construct a diversity-aware core vocabulary with reserved capacity and then check whether sentences contain enough eligible keywords after preprocessing. If a sentence contains no covered content word, the earliest two eligible content words are added to a global refinement queue. If it contains only one covered content word, the earliest uncovered eligible content word is added. For sentences with fewer than two eligible content words after preprocessing, all available content words are retained. If this refinement temporarily enlarges the vocabulary beyond the target budget, low-frequency or semantically redundant words are pruned only when their removal does not reduce any sentence below the two-keyword coverage requirement.

Taken together, the keyword vocabulary turns the original EEG-to-text problem into a fixed keyword-recovery problem whose outputs serve as semantic anchors for reconstruction. It keeps the first-stage target much smaller than a full sentence while preserving enough sentence-level coverage for LLM-based reconstruction. Unless otherwise stated, the primary setting uses the 100-word vocabulary, and the recovered keyword sequences are evaluated as Top-3, Top-5, and Top-7 ordered semantic anchors. This fixed keyword space makes the later comparisons interpretable: Random, Ordered, and Oracle anchors are all evaluated under the same semantic bottleneck.

\subsection{Brain-CLIPLM framework}

\begin{figure}[htbp]
\centering
\includegraphics[width=\textwidth]{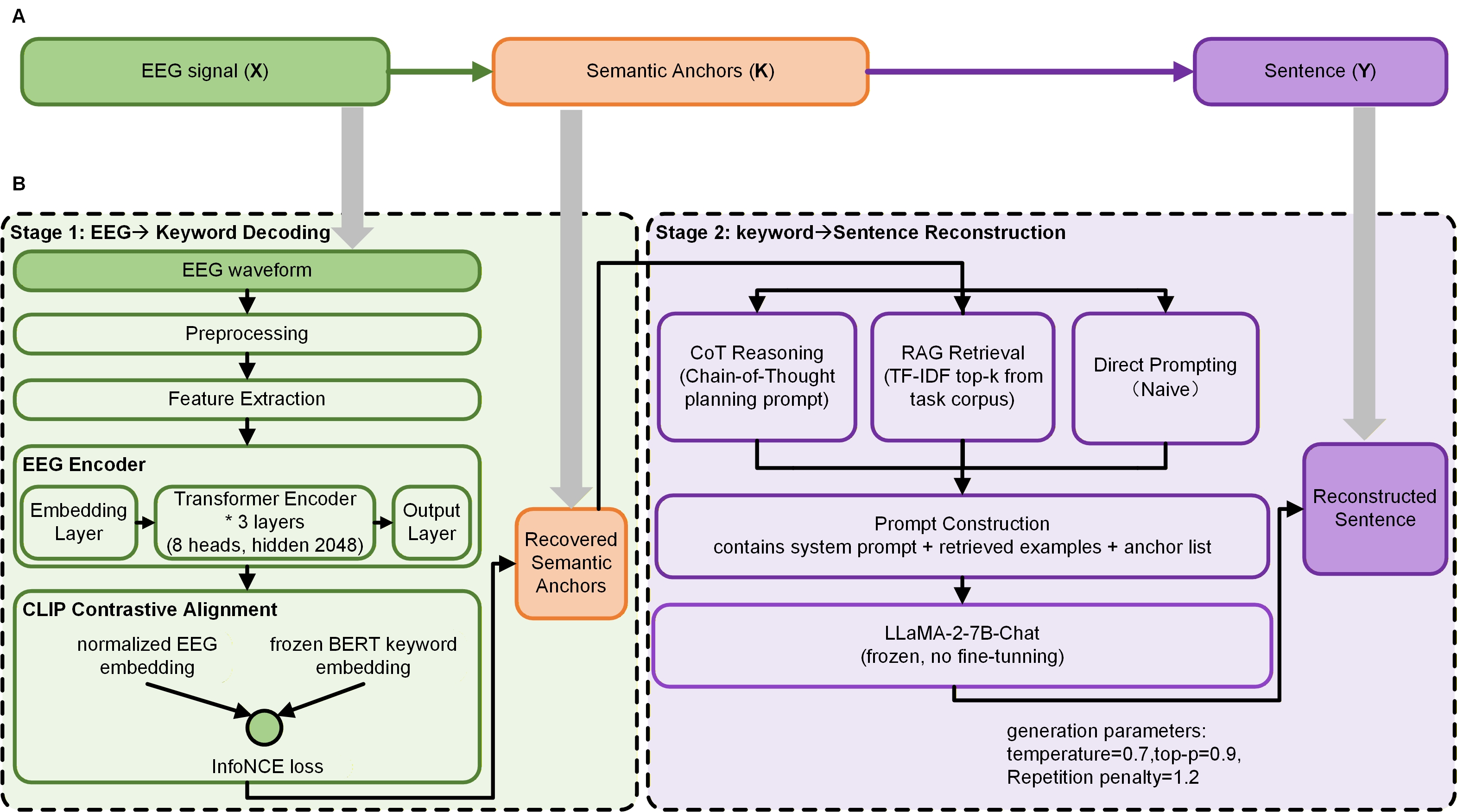}
\caption{Overview of the Brain-CLIPLM two-stage EEG-to-text decoding framework. \textbf{(A)} Conceptual factorization of the decoding process: EEG signals (X) are first mapped to ordered semantic anchors (K), which are then used to reconstruct sentences (Y). \textbf{(B)} Implementation details of the two stages. Stage 1: EEG-to-keyword decoding uses fixation-aligned word-level EEG segments, preprocessing, feature extraction, a Transformer-based EEG encoder, and contrastive alignment with a frozen keyword embedding space to recover semantic anchors. Stage 2: Keyword-to-sentence reconstruction takes the recovered anchors as input to a large language model (LLaMA-2-7B-Chat), guided by chain-of-thought (CoT) reasoning and retrieval-augmented generation (RAG) from the task corpus, producing fluent, sentence-level reconstructions.}
\label{figure3}
\end{figure}

Brain-CLIPLM implements the factorized pathway developed in Section~2: first recover what EEG can more plausibly provide, then reconstruct the sentence from that evidence. As illustrated in Figure~\ref{figure3}, the framework consists of two stages: EEG-to-keyword decoding and keyword-to-sentence reconstruction. In Stage~1, fixation-aligned EEG segments are mapped into a shared semantic embedding space and decoded into an ordered sequence of content keywords from the fixed keyword vocabulary, with the order inherited from the original reading sequence. In Stage~2, this keyword sequence serves as the semantic-anchor scaffold provided to a large language model for sentence reconstruction. This decomposition follows the factorized pathway $X \rightarrow K \rightarrow Y$ (as shown in Figure~\ref{figure3}A), where $X$ denotes EEG evidence, $K$ denotes the recovered semantic-anchor representation, and $Y$ denotes the reconstructed sentence. It separates neural decoding from language generation: EEG is used to recover a compressed semantic scaffold, while the language model expands that scaffold into fluent sentence-level form.

For keyword-to-sentence reconstruction, chain-of-thought (CoT) prompting guides the model to organize the recovered anchors into a structured semantic plan, while retrieval-augmented generation (RAG) retrieves semantically similar sentences from the task-specific corpus as auxiliary references. The final output is therefore not generated from EEG alone, nor from language-model priors alone, but from an explicit semantic scaffold recovered from EEG and expanded through controlled language realization.

\subsubsection{Stage 1: EEG-to-keyword decoding}

Stage~1 (shown in the left panel of Figure~\ref{figure3}B) answers the first part of the central question: can EEG recover the semantic cues needed for sentence reconstruction? Its goal is not to decode every word in the sentence. Instead, it uses fixation-aligned word-level EEG segments to recover an ordered sequence of content keywords that serve as sentence-specific semantic anchors for Stage~2. This design preserves local reading order while avoiding the stronger assumption that scalp EEG can specify every lexical and syntactic decision in the final sentence.

For sentence $i$, let $x_{it}\in\mathbb{R}^{840}$ denote the word-aligned EEG feature vector associated with the $t$-th word position. Following Zheng et al. \citep{ZHENG2026130300}, the EEG encoder uses an MLP projection, learnable positional encodings, a three-layer Transformer encoder with eight attention heads and hidden dimension 2048, and an output MLP that maps each position into a shared 768-dimensional semantic space. On the text side, a frozen BERT-based word-level semantic encoder \citep{devlin2019bert} encodes each candidate keyword independently, yielding a fixed embedding bank for the vocabulary. Thus, each supervised EEG segment is compared with all candidate keywords in the fixed vocabulary.

Training uses a CLIP-style keyword-bank alignment objective \citep{radford2021learning}. For every supervised content-bearing position, the positive target is the keyword corresponding to the lemmatized word at that position when it belongs to the fixed vocabulary, and the remaining vocabulary items serve as semantic negatives. With normalized EEG and keyword embeddings, the alignment loss is written as
\begin{equation}
\mathcal{L}_{\mathrm{align}}
=
-\frac{1}{|\mathcal{B}|}
\sum_{(i,t)\in\mathcal{B}}
\log
\frac{\exp\left(\bar{h}_{it}^{\top}\bar{t}_{y_{it}}/\tau_a\right)}
{\sum_{j=1}^{V}\exp\left(\bar{h}_{it}^{\top}\bar{t}_{j}/\tau_a\right)},
\end{equation}
where $\mathcal{B}$ denotes the supervised EEG positions in a mini-batch, $\bar{h}_{it}$ is the normalized EEG representation, $\bar{t}_j$ is the normalized embedding of keyword $w_j$, $y_{it}$ is the gold keyword index, $V$ is the keyword-vocabulary size, and $\tau_a$ is the temperature. Segments whose aligned words are outside the fixed vocabulary are excluded from this supervised keyword-alignment objective. In implementation, we retain the EEG-side perturbation and discrimination strategy used by Zheng et al. \citep{ZHENG2026130300} as an auxiliary regularization component, but the main cross-modal target remains the fixed keyword bank.

At inference, each valid EEG segment is matched to the most similar keyword in the vocabulary. Candidate keywords are ranked by confidence, and the highest-confidence content-keyword predictions are retained for $m\in\{3,5,7\}$. Their order follows the corresponding word positions in the original reading sequence. If a sentence contains fewer than $m$ valid content keywords, the effective depth is truncated to the number of available keywords. The resulting ordered semantic-anchor sequence is the output of Stage~1. Thus, Stage~1 serves as a semantic compression module: it converts noisy local EEG observations into an ordered anchor scaffold that is more interpretable than raw EEG and less demanding than direct full-sentence decoding.

\subsubsection{Stage 2: Keyword-to-sentence reconstruction}

Stage~2 (shown in the right panel of Figure~\ref{figure3}B) answers the second part of the central question: once EEG has provided an ordered semantic scaffold, can an LLM use it to recover the sentence meaning? We therefore treat sentence recovery as a constrained language-realization problem conditioned on the ordered anchor sequence $\hat{A}_i^{(m)}$ recovered from Stage~1. As in recent EEG-guided language generation frameworks, we use a large language model as the reconstruction module \citep{ZHENG2026130300}, but here it is conditioned on an ordered sequence of content anchors rather than on a full lexical transcript or an unordered keyword set. Stage~1 supplies sentence-specific semantic constraints. Stage~2 restores those parts of the sentence that are unlikely to be directly recoverable from non-invasive EEG, including function words, local compositional relations, and surface fluency. We use LLaMA-2-7B-Chat as the base language model \citep{touvron2023llama} and do not fine-tune it on EEG data so that the roles of neural decoding and sentence realization remain separated.

Operationally, the input to Stage 2 is an ordered anchor list recovered from Stage 1. The anchors are presented to the language model in their original reading order and act as explicit semantic constraints. CoT prompting asks the model to organize these anchors into a sentence-level semantic plan, whereas RAG retrieves semantically similar sentences from the task-specific corpus as auxiliary references. Retrieved sentences provide corpus-grounded examples of possible realizations, but they do not replace the EEG-derived anchors. The output of Stage 2 is one anchor-conditioned sentence reconstruction.

The present task remains sentence-level rather than paragraph-level reconstruction. In the ZuCo benchmark, each sample corresponds to an independent sentence rather than to a discourse segment with genuine preceding or following context. Accordingly, Stage~2 is defined as the recovery of a single standalone sentence from EEG-derived ordered semantic anchors. Retrieved examples, when used, do not provide true discourse context; instead, they serve only as auxiliary sentence-level references drawn from similar sentences in the task-specific corpus. Within this setup, we evaluate four reconstruction settings. In the naive setting, the model receives only the recovered ordered anchor sequence and directly generates one sentence. In the CoT-only setting, we use a structured planning prompt that guides the model to organize the recovered anchors into a coherent sentence hypothesis before generating the final sentence \citep{wei2022chain}. In the RAG-only setting, we retrieve the top-$k=5$ most similar sentences from the full sentence set of the corresponding task and use them as auxiliary references without an explicit reasoning scaffold. In the full CoT+RAG setting, the model first identifies the retrieved sentences most compatible with the recovered anchor sequence, then uses the shared semantic structure between the anchors and the retrieved candidates to form a sentence plan, and finally generates one sentence. This organization allows the contribution of reasoning and retrieval grounding to be examined under the same anchor input.

For retrieval-augmented settings, retrieval is performed over the full sentence set of the corresponding task using TF-IDF similarity over content words \citep{salton1988term, lewis2020retrieval}. In other words, the candidate pool is the task-specific sentence library itself rather than a small pre-filtered subset, and the top-$k$ retrieved sentences are then provided to the language model as auxiliary references. In all reconstruction settings, the ordered anchors are presented to the language model as a single anchor list in sentence order. The prompts are designed to enforce three common principles throughout: anchor fidelity, minimal unsupported additions, and single-sentence output. Generation parameters are fixed throughout the experiments at temperature $=0.7$, top-$p=0.9$, repetition penalty $=1.2$, and maximum output length of 100 tokens. Taken together, Stage~2 expands the compressed anchor scaffold into a fluent sentence while keeping reconstruction tied to sentence-specific semantic evidence. The language model is therefore used to realize and constrain meaning, not to replace the neural decoding stage.

Taken together, the two stages implement the factorized pathway proposed in Section~2. Stage~1 estimates an ordered semantic representation from EEG, and Stage~2 reconstructs a sentence from that representation using retrieval and reasoning prompts. This division of labor is the main methodological answer to the central question of the paper: EEG is not treated as a direct source of complete sentence form, but as a source of compressed semantic anchors that can guide constrained sentence recovery.

\subsection{Evaluation protocol and statistical analysis}

The evaluation is designed to test the same question that motivates Brain-CLIPLM: whether EEG is better used to recover semantic anchors before sentence reconstruction, rather than to directly specify a full sentence. We therefore evaluate Stage~1 and Stage~2 separately and then examine their combined effect. Two benchmark settings are used for different reporting purposes. Descriptive retrieval performance, including Top-5, Top-10, and Top-25 accuracies, is reported on the 30-participant multi-subject benchmark (SR v1.0 + NR v1.0 + NR v2.0), which provides the broadest participant coverage. Repeated-measures inferential tests are reported on the 12-participant matched-material benchmark (SR v1.0 + NR v1.0 + TSR v1.0). This matched-material setting keeps the participant cohort and task/material configuration fixed across the compared conditions, thereby reducing the possibility that differences among target granularities or reconstruction strategies are confounded by benchmark composition. Each reported F-test is explicitly labeled with the benchmark setting and exact N on which it was computed.

For Stage~1, we assess whether the recovered Top-3, Top-5, and Top-7 ordered anchors are grounded in the source sentence after the same lowercasing and lemmatization rules used in keyword vocabulary construction. AnchorHit@$m$ measures the proportion of recovered anchors that also occur in the source sentence, whereas SentenceAnchorAll@$m$ counts a sentence as correct only when all recovered anchors are grounded in the source sentence. Because no manual keyword labels are available, these measures are intentionally conservative: they count only anchors that literally appear in the normalized source sentence. At the same time, Stage~1 is not judged only by literal anchor overlap. Its main role is to provide ordered semantic cues for Stage~2, so we also evaluate its downstream utility through the resulting sentence retrieval performance.

For Stage~2, the primary endpoint is sentence retrieval accuracy. Given a reconstructed sentence, all unique sentences in the corresponding task subset are ranked by cosine similarity between Sentence-BERT embeddings \citep{reimers2019sentence}. Reconstruction is counted as successful when the ground-truth sentence appears within the Top-5, Top-10, Top-15, Top-20, or Top-25 retrieved candidates. The retrieval pool is the full sentence inventory of the corresponding task subset: 400 sentences for SR v1.0, 300 for NR v1.0, 349 for NR v2.0, and 407 for TSR v1.0. Chance level is computed from the corresponding task-specific pool size. We use retrieval accuracy as the primary sentence-level endpoint because Brain-CLIPLM aims to recover sentence meaning from EEG-derived ordered anchors rather than to reproduce surface wording exactly. For comparability with prior EEG-to-text studies, we also report BLEU-1, BLEU-2, BLEU-3, ROUGE-1 F1, and BERTScore F1 as secondary text-generation metrics \citep{papineni2002bleu, lin2004rouge, zhang2019bertscore, ZHENG2026130300}. These secondary metrics provide useful reference points for lexical overlap and semantic similarity, but they are not treated as substitutes for the retrieval-based endpoint. The sentence-level evaluation is organized around two groups of comparisons. The first group is the main ablation study, including Naive generation, RAG only, CoT only, and CoT+RAG (Ours), which tests how retrieval and reasoning contribute under the same ordered-anchor input. The second group contains the anchor-quality controls: Random anchors, Ordered anchors (Ours), and Oracle anchors. Random anchors preserve the reconstruction pipeline but remove sentence-specific EEG--anchor correspondence; Ordered anchors use the Stage~1 recovered anchors; and Oracle anchors provide an approximate upper bound using ground-truth anchors from the source sentence. These Random--Ordered--Oracle anchor controls are reported as descriptive control-baseline comparisons rather than permutation-based inferential tests. Together with the target-granularity comparison across 3, 5, and 7 anchors, these analyses test whether sentence recovery depends on EEG-derived ordered semantic evidence rather than on language-model priors alone.

All statistical tests were two-tailed. Summary statistics are reported as participant-level mean $\pm$ SD in the corresponding evaluation setting. For each participant, retrieval accuracy is first averaged over held-out test sentences, and the reported mean and standard deviation are then computed across the participants in the corresponding benchmark. Thus, the participant, rather than the sentence, is treated as the unit of summary and inference. For inferential comparisons, we used repeated-measures ANOVA for multi-condition comparisons and paired $t$-tests for pairwise comparisons, with Bonferroni correction applied where appropriate. Statistical significance was assessed at $p < 0.05$ unless otherwise stated. 

\section{Results}

The Results section follows the same causal chain as the Brain-CLIPLM pipeline. We first test whether semantic anchors can be recovered from EEG. We then ask which anchor granularity is best matched to sentence-level reconstruction, whether those anchors support sentence recovery, and whether the observed gains depend on sentence-specific neural information rather than language-model priors alone. Descriptive retrieval values in the main Results are reported on the 30-participant multi-subject benchmark (SR v1.0 + NR v1.0 + NR v2.0), which provides the broadest participant coverage. Repeated-measures ANOVAs are reported as matched-material inferential analyses computed on the 12-participant benchmark (SR v1.0 + NR v1.0 + TSR v1.0). Each reported F-test is labeled with its benchmark setting and exact sample size. This distinction explains the degrees of freedom of the reported inferential statistics and does not change the interpretation of the retrieval results. Thus, the section is organized around the same question that motivates the method: whether EEG should first be used to recover semantic anchors and only then be used for sentence-level reconstruction. Sentence retrieval is the primary endpoint of this study because Brain-CLIPLM aims to recover sentence-level meaning from EEG-derived semantic anchors; secondary generation metrics (BLEU, ROUGE, BERTScore) are reported for comparability with prior work.

\subsection{Decoding semantic anchors from EEG}

\begin{figure}[htbp]
\centering
\includegraphics[width=\textwidth]{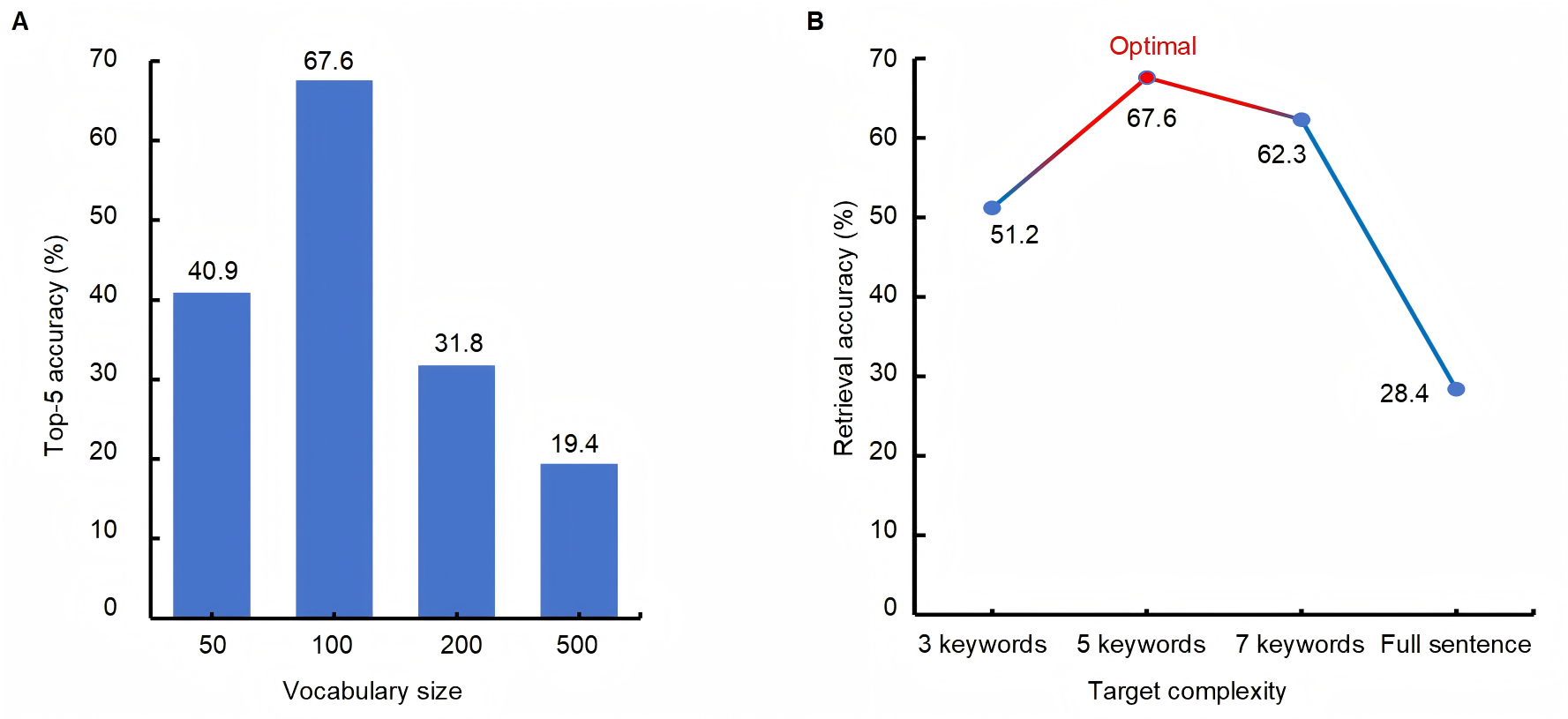}
\caption{Stage 1 keyword retrieval and target-granularity analyses. \textbf{(A)} Keyword retrieval accuracy across vocabulary sizes of 50, 100, 200, and 500 words. Bars show Top-5 accuracy, and the corresponding Top-10 and Top-25 results for the primary 100-word setting are reported in the text. \textbf{(B)} Target-complexity scan across 3 keywords, 5 keywords, 7 keywords and full-sentence targets. Performance peaks at an intermediate granularity, consistent with the granularity-matching principle.}
\label{figure4}
\end{figure}

The first empirical question is whether the intermediate anchor code can be recovered from EEG at all. Because Brain-CLIPLM first builds a fixed keyword vocabulary and then aligns word-aligned EEG evidence to that vocabulary, Stage~1 must be evaluated before the anchor representation can be treated as a useful interface for sentence reconstruction. This analysis is the central neural decoding test of the framework, because it asks whether sentence-specific semantic cues can be recovered from EEG before any LLM-based sentence realization is applied. Accordingly, Stage~1 performance serves as the principal neurocomputational evidence for the semantic compression pathway, because it evaluates EEG-derived semantic-anchor recovery independently of the downstream sentence-reconstruction module. Across vocabulary sizes of 50, 100, 200, and 500 words, Top-5 accuracy is 40.9\%, 67.6\%, 31.8\%, and 19.4\%, respectively (Figure~\ref{figure4}A). These results indicate that anchor recovery is neither trivial nor prohibitively weak: it remains a difficult decoding problem, but one that is substantially more tractable than full-sentence recovery. We therefore use the 100-word vocabulary as the primary setting in the subsequent analyses. Recoverability at the single-subject level is only part of the story. A useful intermediate code should also generalize beyond subject-specific calibration. We therefore test whether the recovered semantic-anchor signal retains structure under leave-one-subject-out evaluation, where the model must rely on what is shared across readers rather than on idiosyncratic within-subject patterns. Under leave-one-subject-out cross-subject validation (using the single-subject task combination SR v1.0 + NR v1.0 + TSR v1.0 across 12 participants), mean Top-5 accuracy reaches 31.7\% (SD = 3.2\%). Relative to single-subject accuracy, cross-subject decoding preserves part of the recoverable signal, indicating that the anchor code contains a substantial shared component across participants. Taken together, these results show that semantic anchors are not merely a theoretical convenience. They define a recoverable intermediate target that preserves sentence-relevant information under both subject-specific and cross-subject EEG decoding settings, providing the necessary bridge to sentence-level reconstruction.

\subsection{Identifying the target scale best matched to EEG}

Having established that semantic anchors are recoverable from EEG, we next ask how much anchor information should be passed to the reconstruction stage. If sentence-level text is too rich a target for non-invasive EEG, then performance should not improve simply by asking the decoder to recover increasingly detailed outputs; it should instead peak at an intermediate semantic scale. We therefore compare sentence retrieval across four target settings---3 keywords, 5 keywords, 7 keywords, and full-sentence direct decoding with a Transformer encoder--decoder. Three keywords achieve 51.2\% Top-5 accuracy, five keywords achieve 67.6\%, seven keywords drop to 62.3\%, and full-sentence direct decoding reaches only 28.4\%, significantly worse than any keyword-based reconstruction (all \(p<0.05\)), as shown in Table~\ref{table1} and Figure~\ref{figure4}B. For the matched-material inferential analysis (SR v1.0 + NR v1.0 + TSR v1.0, N = 12), a one-way repeated-measures ANOVA confirmed a significant main effect of target granularity (\(F(3,33)=34.2\), \(p<0.05\)). Post hoc tests show that five keywords outperform all other conditions (all \(p<0.01\)). This pattern is consistent with the granularity-matching principle: under the present setup, EEG appears to contain enough information to constrain a compact semantic description, but not enough to support direct recovery of a full sentence. These results answer the target-scale question. The best-performing target is not the most detailed linguistic output, but a compact anchor representation that preserves sentence-relevant meaning while avoiding the full burden of sentence-form recovery. This result establishes the target scale used in the remainder of the experiments.

\subsection{Reconstructing sentences from decoded anchors}

Once anchor recovery is established, the next question is whether those anchors can support sentence reconstruction at the endpoint that matters operationally. This subsection therefore evaluates whether the recovered anchor scaffold can be converted into sentence-level semantic recovery, rather than remaining a local keyword-matching result.

\begin{figure}[htbp]
\centering
\includegraphics[width=\textwidth]{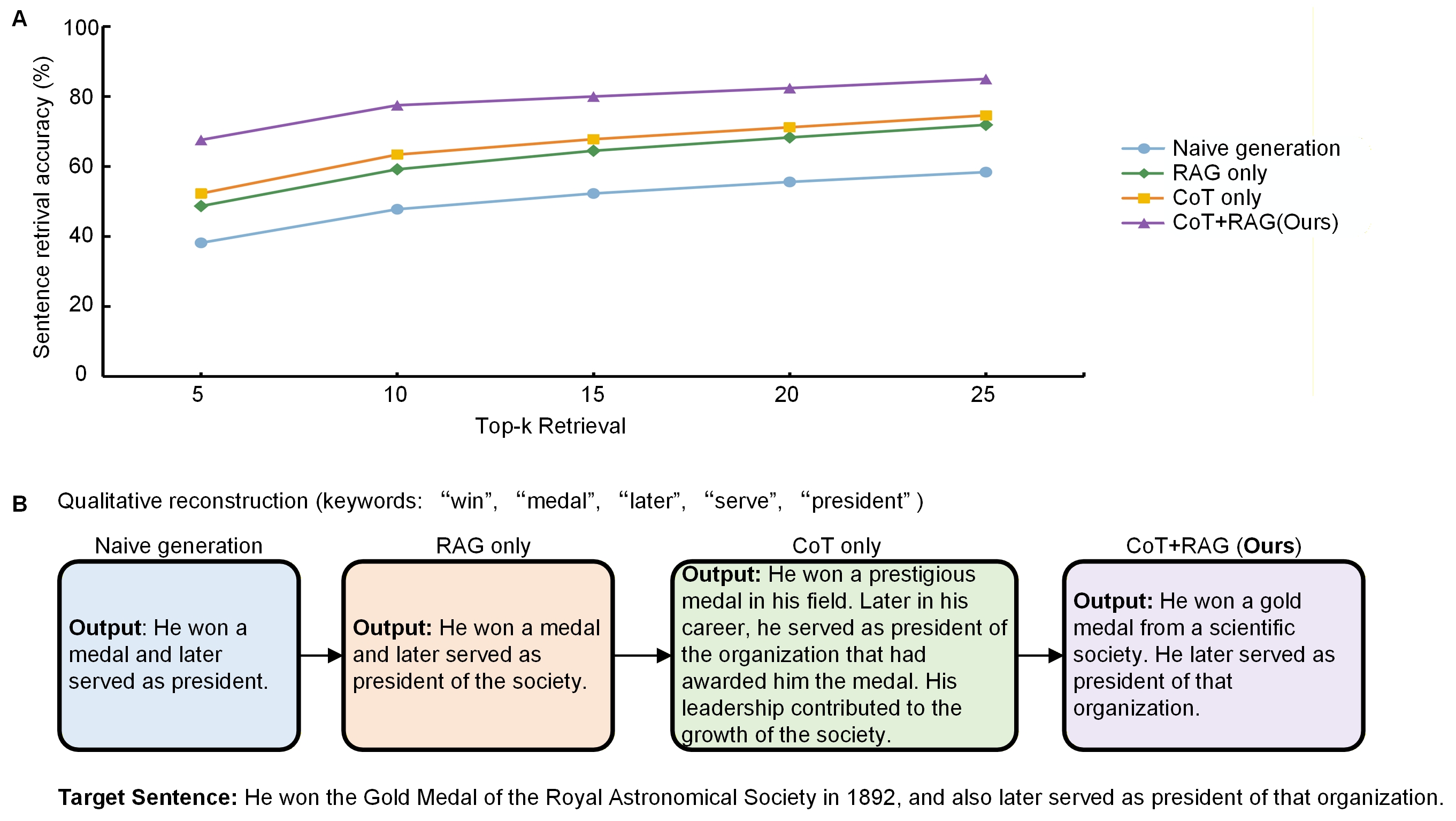}
\caption{Sentence recovery performance and qualitative reconstruction examples. \textbf{(A)} Top-$k$ sentence retrieval accuracy across reconstruction strategies. The main ablation study shows that CoT+RAG (Ours) achieves the highest retrieval under this setup, indicating complementary contributions from reasoning and retrieval grounding. \textbf{(B)} Representative qualitative example comparing Naive generation, RAG only, CoT only and CoT+RAG (Ours) for the same recovered keyword set.}
\label{figure5}
\end{figure}

Performance differs markedly across reconstruction strategies as illustrated in Table~\ref{table2} and Figure~\ref{figure5}A. CoT+RAG (Ours) reaches 67.6\% Top-5 accuracy and 85.0\% Top-25 accuracy. These results show that the recovered anchors carry enough sentence-level information to support constrained reconstruction, not merely local keyword matching. CoT alone reaches 52.3\%, RAG alone 48.7\%, and their combination adds 15--19 points beyond either alone. For the matched-material inferential analysis (SR v1.0 + NR v1.0 + TSR v1.0, N = 12), a two-way repeated-measures ANOVA identified significant main effects of CoT (\(F(1,11)=42.6\), \(p<0.05\)) and RAG (\(F(1,11)=38.2\), \(p<0.05\)), with a significant interaction (\(F(1,11)=8.7\), \(p<0.05\)). The full pipeline therefore gains not simply from adding more components, but from combining complementary forms of structure: anchor-guided reasoning and retrieval-grounded specificity. 

Table~\ref{table3} reports literature baseline values for EEG-to-Text and Guiding LLMs together with the present results on the shared text-generation metrics. Among these references, the comparison with Guiding LLMs is the closest protocol-level benchmark because the present study follows Zheng et al. (2026) in benchmark construction, task combinations, data partitioning, sample filtering, 840-dimensional EEG feature construction, and the Stage~1 EEG encoder design. The text side, however, is intentionally different: Brain-CLIPLM uses a \citet{zhou2024towards}-style frozen word-level semantic encoder adapted to a fixed anchor bank, rather than the contextual token-semantic encoder used in Guiding LLMs. Table~\ref{table3} should therefore be read as a strong protocol-near comparison rather than as a fully identical implementation rerun. Under this comparison, Brain-CLIPLM yields higher values than the EEG-to-Text baseline \citet{wang2022open} and Guiding LLMs framework \citet{ZHENG2026130300} on BLEU-1, BLEU-2, BLEU-3, ROUGE-1 F1, and BERTScore. We use these comparisons to position Brain-CLIPLM within the broader literature, while retaining the within-study target-complexity tests, pathway comparisons, and controls as the main evidential basis for the semantic compression account.

Taken together, the sentence-level results show that recovered anchors can support constrained sentence recovery, especially when reasoning and retrieval are used to convert the anchor scaffold into corpus-grounded sentence candidates. The comparison with prior work provides useful context, but the central evidence remains the within-study pattern: sentence reconstruction improves when EEG is first translated into ordered semantic anchors.

\subsection{Control analyses: neural information versus language-model priors}

The final result addresses the strongest alternative explanation for the preceding findings. Because the reconstruction stage uses a language model and a task-specific sentence corpus, fluent outputs could in principle be driven mainly by language priors, with EEG-derived anchors contributing little sentence-specific information. 

The control analyses evaluate this possibility through a descriptive Random--Ordered--Oracle control-baseline comparison. Table~\ref{table4} compares three conditions: random anchors, ordered anchors (ours), and oracle anchors. The ordered anchors (Ours) setting, instantiated here with the full CoT+RAG reconstruction pipeline, reaches 67.6\%, remaining below the oracle-anchor upper bound. The gap between ordered anchors and oracle anchors is much smaller than that between random anchors and ordered anchors. Qualitative examples make the same point visible at the sentence level. As shown in Figure~\ref{figure5}B, for the decoded keywords [``win'', ``medal'', ``later'', ``serve'', ``president''], Naive generation produces a fluent but simple sentence; CoT only generates a more detailed and plausible sentence; RAG only introduces corpus-grounded specificity but remains limited by lexical overlap; and CoT+RAG (Ours) combines both strengths to produce a sentence closest to the ground truth. The example therefore mirrors the quantitative pattern: Ordered anchors (Ours) provide sentence-specific semantic constraints, CoT organizes them into a coherent plan, and retrieval contributes corpus-grounded specificity. Together, the controls and qualitative examples support the conclusion that the gains of the full pipeline are driven by neural information that the language model can use, rather than by language priors acting on their own. This final analysis closes the main evidential gap: Brain-CLIPLM improves because the intermediate anchors carry sentence-specific EEG-derived evidence, not simply because an LLM can generate plausible text.

Overall, the above results close the loop opened in the Introduction. The anchor-decoding analyses first show that this intermediate representation is recoverable from EEG. The target-granularity analyses then show why a compact anchor target is better matched to EEG than direct full-sentence decoding. The reconstruction analyses show that recovered anchors can support sentence-level semantic recovery. Finally, the control analyses show that the observed gains depend on EEG-derived ordered anchors rather than language-model priors alone. Across these results, the most consistent pattern is that EEG-derived ordered anchors provide a more appropriate intermediate target than direct full-sentence decoding, supporting the practical value of recovering semantic anchors first and reconstructing fluent sentences second.

\section{Discussion}

This study starts from a simple question: why should EEG directly decode a complete sentence if sentence meaning can often be supported by a small number of content-bearing cues? The results support the corresponding answer. Under non-invasive EEG constraints, a compact ordered anchor representation is a more tractable and interpretable intermediate target than full sentence form. We discuss this conclusion in three steps: the decoding target, the role of the two-stage framework, and the limits of the present evidence.

\subsection{The target of non-invasive EEG decoding}

The main contribution of Brain-CLIPLM is a change in target definition. Instead of treating EEG as a direct source sequence for sentence generation, the framework treats EEG as partial semantic evidence. Stage~1 asks what content-bearing anchors can be recovered from the neural signal; Stage~2 asks how those anchors can be expanded into a sentence. Thus, Stage~1 provides the primary evidence for EEG-based semantic-anchor recovery, whereas Stage~2 tests whether the recovered anchors can support constrained sentence-level reconstruction. This distinction is important because it prevents the language model from hiding the weakness of the neural signal behind fluent output.

The empirical pattern is consistent with this reframing. Performance does not improve monotonically as the target becomes richer. It peaks at an intermediate anchor granularity, whereas direct full-sentence decoding performs substantially worse. The pathway comparison leads to the same conclusion: random anchors provide little useful constraint, oracle anchors provide a high upper bound, and EEG-derived ordered anchors fall between them. This is the pattern expected if EEG contains partial but meaningful semantic information rather than enough information to specify the whole sentence directly.

\subsection{The role of the two-stage framework}

The two-stage framework is useful because it assigns distinct roles to the neural decoder and the language model. The neural decoder is responsible for recovering sentence-specific semantic anchors from EEG, whereas the language model restores function words, local composition, and surface fluency around those anchors. This division of labor matches the central motivation of the paper: word-level semantic cues are easier to recover from non-invasive EEG than complete sentence form, and sentence-level language can be reconstructed more plausibly once those cues are available. In this sense, Brain-CLIPLM does not treat the language model as a substitute for neural decoding. It uses the language model to realize and constrain meaning from an explicit semantic scaffold recovered from EEG.

The control analyses show why this separation matters. When the anchor sequence is randomized, the same reconstruction pipeline can still produce grammatical sentences, but retrieval accuracy drops sharply. This indicates that fluency alone is not enough: the full system performs well only when the LLM receives ordered anchors that carry sentence-specific information from EEG. This also explains why sentence retrieval is the primary endpoint. The present study does not claim unrestricted open-vocabulary generation from EEG; it asks whether sentence-level meaning can be recovered within a fixed candidate space from EEG-derived semantic anchors. BLEU, ROUGE, and BERTScore remain useful secondary descriptions, but retrieval accuracy more directly tests whether the reconstructed sentence points back to the correct sentence-level meaning.

The relationship between the semantic compression hypothesis and Brain-CLIPLM is therefore not simply that of a motivation and a model. The hypothesis defines the representational target that non-invasive EEG is expected to support: a compact, ordered semantic-anchor representation rather than the full lexical--syntactic form of a sentence. Brain-CLIPLM provides a concrete way to test this target. By decomposing decoding into EEG-to-keyword recovery followed by keyword-to-sentence reconstruction, the framework turns the hypothesis into an evaluable pathway. The results do not prove that the brain stores sentences as keyword lists, nor do they imply that semantic anchors uniquely determine a sentence. Rather, they provide evidence that EEG-derived semantic anchors offer a more tractable and useful intermediate representation for constrained sentence recovery than direct full-sentence decoding.

Taken together, the findings can be understood as the convergence of three perspectives: semantic compression, structural organization, and information-scale constraints. The semantic compression account explains why exact recovery of every token is unnecessary for sentence-level meaning. The structural perspective explains why one anchor is too coarse, because sentence meaning is distributed over multiple coherent units rather than a single point. The information-scale perspective explains why direct sentence decoding is too demanding for non-invasive EEG. These perspectives converge on the same intermediate target: a small ordered set of semantic anchors. This convergence also clarifies the relationship between Brain-CLIPLM and recent work on constituent-constrained prediction. That work supports the broader view that language comprehension relies on compressed, structured representations rather than uniform token-by-token prediction \citep{zou2026constituent}. Brain-CLIPLM does not attempt to decode parser-defined constituents, and semantic anchors are not treated as syntactic boundary markers. Instead, the constituent evidence motivates a weaker but important principle: useful language representations are often multi-unit and compressed. Brain-CLIPLM instantiates this principle at the level of EEG decoding, where the intermediate representation must also remain recoverable from noisy neural data.

The theoretical implication is that the advantage of Brain-CLIPLM does not arise from any single component alone. It comes from aligning the decoding target with both the organization of language and the constraints of neural measurement. EEG supplies a compact semantic scaffold, and the language model converts that scaffold into a sentence-level interpretation. This separation suggests that future EEG-to-language systems may benefit less from forcing increasingly complex sentence-level decoders onto noisy signals, and more from identifying intermediate representations that are simultaneously structured, compressed, and neurally recoverable.

\subsection{Limitations and future directions}

Several limitations bound the claims. First, Brain-CLIPLM is evaluated in a constrained sentence-recovery setting, not in fully open-vocabulary generation of previously unseen sentences from EEG \citep{wang2022open, liu2024eeg2text}. Retrieval over a task-specific sentence pool is used as a controlled test of semantic recovery, not as evidence that unrestricted EEG-to-text generation has been solved. Second, the benchmark is based on fixation-aligned reading data in ZuCo. The present neural decoding component should therefore be interpreted as supervised, fixation-aligned EEG-to-keyword or semantic-anchor recovery, not as blind segmentation of continuous EEG into word-level units. Fixation durations and segment boundaries are part of the measurement protocol, and future work should examine how segment-duration variability, sliding-window segmentation, and full-sentence EEG windows affect anchor recovery. The same factorized pathway still needs to be tested in inner speech, imagined language, and more spontaneous communicative settings \citep{nieto2022thinking, kunz2025inner}. Third, the keyword vocabulary is fixed and restricted to high-frequency content words. This makes the task learnable and interpretable, but it also means that the semantic compression hypothesis is tested here in a finite-vocabulary form. Fourth, the current feature protocol follows the standard ZuCo/Zheng-style eight-band representation and does not include delta-band features. Delta-inclusive representations and longer temporal windows may be better suited to capturing long-timescale semantic integration and should be evaluated in future work.

Future work should therefore extend the present framework in two directions. Methodologically, richer anchor vocabularies, adaptive anchor selection, delta-inclusive feature construction, eye-tracking-free segmentation, and direct estimates of neural semantic information could provide stronger tests of the semantic compression account. Translationally, anchor-based decoding should be evaluated in settings closer to practical communication BCIs, where the goal is often not verbatim transcription of every token but reliable recovery of intended meaning.

\section{Conclusion}

Within the constrained ZuCo sentence pool and fixed keyword-vocabulary settings, the present study examines whether non-invasive EEG should be used to directly decode complete sentences or to first recover a smaller set of semantic anchors. Brain-CLIPLM follows the second route. It decomposes EEG-to-text decoding into ordered semantic-anchor recovery and LLM-based sentence reconstruction. The results support this factorized formulation. Across the ZuCo benchmark, intermediate anchor granularity outperforms both overly sparse anchors and direct full-sentence decoding. EEG-derived ordered anchors improve constrained sentence recovery relative to random anchors and remain below oracle anchors, indicating that they carry partial but useful sentence-specific information. 

Overall, Brain-CLIPLM suggests that current non-invasive EEG-to-text decoding may be more effective when it first recovers gist-level semantic structure and then uses language models to realize that structure as fluent text. The central conclusion is therefore bounded but clear: EEG does not need to specify every word of a sentence to support constrained sentence-level recovery within a predefined candidate space; it can first provide semantic anchors that make reconstruction tractable. More broadly, this factorized view offers a principled direction for future EEG-to-language systems: identify intermediate representations that are compact enough to be recoverable from noisy neural signals, yet structured enough to support meaningful sentence reconstruction.

\section*{Data Availability Statement}
The Zurich Cognitive Language Processing Corpus (ZuCo) benchmark used in this study is constructed from the publicly available ZuCo 1.0 and ZuCo 2.0 datasets. The two datasets are available here: \url{https://osf.io/q3zws/} and \url{https://osf.io/2urht/}. 

\section*{Author Contributions}
XY: Conceptualization, Methodology, Investigation, Formal analysis, Visualization, Writing—original draft. HT: Formal analysis, Writing—review \& editing. YL: Conceptualization, Writing—review \& editing. JZ: Software, Writing—review \& editing. SL: Supervision, Validation, Funding acquisition, Writing—review \& editing. GP: Supervision. All authors contributed to the article and approved the submitted version.

\section*{Funding}
This work is supported by the STI 2030 Major Projects (2021ZD0200403) and the Zhejiang Provincial Natural Science Foundation of China (Grant No. LD24F030002).

\section*{Conflict of Interest Statement}
The authors declare that the research was conducted in the absence of any commercial or financial relationships that could be construed as a potential conflict of interest.

\section*{Generative AI statement}
No generative AI tools were used to draft, edit, or prepare the manuscript text. LLaMA-2-7B-Chat was used only as an experimental component of the Stage~2 sentence-reconstruction module described in the Methods, and not as a manuscript-preparation tool.

\bibliographystyle{Frontiers-Harvard}
\bibliography{test}

\clearpage
\section*{Tables}

\nolinenumbers
\begingroup
\renewcommand{\arraystretch}{1.5}
{\fontsize{10}{12}\selectfont
\arrayrulecolor{gray!60}
\setlength{\tabcolsep}{4pt}

\captionof{table}{Sentence retrieval across target granularities. Best performance is obtained with five keywords.}
\label{table1}
\centering
\begin{tabularx}{\textwidth}{|>{\raggedright\arraybackslash}p{4cm}|Y|Y|Y|}
\hline
\rowcolor{gray!200} \textcolor{white}{\textbf{Target granularity}} & \textcolor{white}{\textbf{Top-5}} & \textcolor{white}{\textbf{Top-10}} & \textcolor{white}{\textbf{Top-25}} \\
\hline
3 keywords & 51.2\% $\pm$ 3.5\% & 62.1\% $\pm$ 4.1\% & 72.3\% $\pm$ 4.7\% \\
\hline
\textbf{5 keywords} & \textbf{67.6\% $\pm$ 4.2\%} & \textbf{77.5\% $\pm$ 4.5\%} & \textbf{85.0\% $\pm$ 3.9\%} \\
\hline
7 keywords & 62.3\% $\pm$ 3.8\% & 73.8\% $\pm$ 4.3\% & 81.2\% $\pm$ 5.0\% \\
\hline
Full sentence & 28.4\% $\pm$ 3.1\% & 35.2\% $\pm$ 3.6\% & 43.6\% $\pm$ 4.2\% \\
\hline
\end{tabularx}

\vspace{12pt}

\captionof{table}{Sentence retrieval across reconstruction strategies. The main ablation study shows that CoT+RAG (Ours) gives the strongest retrieval performance.}
\label{table2}
\centering
\begin{tabularx}{\textwidth}{|>{\raggedright\arraybackslash}p{4cm}|Y|Y|Y|}
\hline
\rowcolor{gray!200} \textcolor{white}{\textbf{Condition}} & \textcolor{white}{\textbf{Top-5}} & \textcolor{white}{\textbf{Top-10}} & \textcolor{white}{\textbf{Top-25}} \\
\hline
Naive generation & 38.2\% $\pm$ 3.3\% & 47.8\% $\pm$ 3.9\% & 58.4\% $\pm$ 4.3\% \\
\hline
RAG only & 48.7\% $\pm$ 3.7\% & 59.2\% $\pm$ 4.2\% & 71.9\% $\pm$ 4.4\% \\
\hline
CoT only & 52.3\% $\pm$ 3.9\% & 63.4\% $\pm$ 4.3\% & 74.6\% $\pm$ 4.8\% \\
\hline
\textbf{CoT+RAG (Ours)} & \textbf{67.6\% $\pm$ 4.2\%} & \textbf{77.5\% $\pm$ 4.5\%} & \textbf{85.0\% $\pm$ 3.9\%} \\
\hline
\end{tabularx}

\vspace{12pt}

\captionof{table}{Comparison with literature baselines on shared text-generation metrics. The Guiding LLMs row provides the closest protocol-near reference, aligning benchmark construction, task combinations, data partitioning, sample filtering, 840-dimensional EEG features, and the Zheng-style EEG encoder, whereas Brain-CLIPLM differs on the text side by using a Zhou et al. (2024)-style frozen word-level semantic encoder adapted to a fixed anchor bank.}
\label{table3}
\centering
\begin{tabularx}{\textwidth}{|>{\raggedright\arraybackslash}p{4.9cm}|Y|Y|Y|Y|Y|}
\hline
\rowcolor{gray!200} \textcolor{white}{\textbf{Method}} & \textcolor{white}{\textbf{BLEU-1}} & \textcolor{white}{\textbf{BLEU-2}} & \textcolor{white}{\textbf{BLEU-3}} & \textcolor{white}{\textbf{ROUGE-1 F1}} & \textcolor{white}{\textbf{BERTScore}} \\
\hline
EEG-to-Text \citep{wang2022open} & 17.0 & 9.3 & 4.9 & 17.8 & 83.3 \\
\hline
Guiding LLMs \citep{ZHENG2026130300} & 21.3 & 11.9 & 7.7 & 20.3 & 84.8 \\
\hline
\textbf{Brain-CLIPLM (Ours)} & \textbf{22.4} & \textbf{12.4} & \textbf{8.1} & \textbf{22.9} & \textbf{86.2} \\
\hline
\end{tabularx}

\vspace{12pt}

\captionof{table}{Control-baseline comparison of anchor conditions. Performance increases from Random anchors to Ordered anchors (Ours) and reaches its highest level under Oracle anchors.}
\label{table4}
\centering
\begin{tabularx}{\textwidth}{|>{\raggedright\arraybackslash}p{4cm}|Y|Y|Y|}
\hline
\rowcolor{gray!200} \textcolor{white}{\textbf{Condition}} & \textcolor{white}{\textbf{Top-5 accuracy}} & \textcolor{white}{\textbf{$\Delta$ from oracle}} & \textcolor{white}{\textbf{Recovery relative to oracle}} \\
\hline
Random anchors & 12.3\% $\pm$ 2.9\% & -78.9\% & 13.5\% \\
\hline
\textbf{Ordered anchors (Ours)} & \textbf{67.6\% $\pm$ 4.2\%} & \textbf{-23.6\%} & \textbf{74.1\%} \\
\hline
Oracle anchors & 91.2\% $\pm$ 3.3\% & 0\% & 100\% \\
\hline
\end{tabularx}

}

\endgroup

\clearpage
\section*{Figure captions}

\clearpage
\end{document}